\documentclass[conference]{IEEEtran}
\AtEndDocument{\par\leavevmode}

\usepackage{courier}

\usepackage{graphicx}
\ifCLASSINFOpdf
\else
\fi
\usepackage{amsmath}
\usepackage{amsfonts}
\usepackage{amsthm}
\ifCLASSOPTIONcompsoc
  \usepackage[caption=false,font=normalsize,labelfont=sf,textfont=sf]{subfig}
\else
  \usepackage[caption=false,font=footnotesize]{subfig}
\fi
\usepackage{url}

\usepackage{widetext}

\newcommand{\bs}{\mathbf s}
\newcommand{\del}{\partial}
\newcommand{\atanh}{{\rm arctanh}}
\newcommand{\pr}{{\rm Pr}}

\newtheorem{theorem}{Theorem}

\begin{document}
\bstctlcite{IEEEexample:BSTcontrol}

%
\title{An Analytic Solution to the Inverse Ising Problem in the Tree-reweighted Approximation}

\IEEEoverridecommandlockouts
\author{
  \IEEEauthorblockN{Takashi Sano$^\dagger$\thanks{$\dagger$ Current affiliation: Seikei University, Tokyo, Japan, 180--8633}}
  \IEEEauthorblockA{%
    National Institute of Advanced Industrial Science and Technology,\\
    Artificial Intelligence Research Center,\\
Tokyo, Japan, 135--0064\\
Email: tsano@st.seikei.ac.jp}
}


%


\IEEEoverridecommandlockouts
\IEEEpubid{\makebox[\columnwidth]{    
\copyright2018
IEEE \hfill} \hspace{\columnsep}\makebox[\columnwidth]{ }}

\maketitle

\begin{abstract}
  Many iterative and non-iterative methods have been developed for
  inverse problems associated with Ising models.
  Aiming to derive an accurate non-iterative method for the inverse problems,
  we employ the tree-reweighted approximation.
  Using the tree-reweighted approximation,
  we can optimize the rigorous lower bound of the objective function.
  By solving the moment-matching and self-consistency conditions analytically,
  we can derive the interaction matrix
  as a function of the given data statistics.
  With this solution, we can obtain the optimal interaction matrix without iterative computation.
  To evaluate the accuracy of the proposed inverse formula,
  we compared our results to those obtained by
  existing inverse formulae derived with other approximations.
  In an experiment to reconstruct the interaction matrix,
  we found that the proposed formula returns the best estimates
  in strongly-attractive regions for various graph structures.
  We also performed an experiment using real-world biological data.
  When applied to finding the connectivity of neurons from spike train data,
  the proposed formula gave the closest result to that
  obtained by a gradient ascent algorithm,
  which typically requires thousands of iterations.
\end{abstract}

\section{Introduction}
Probabilistic graphical models
are intriguing because they 
are employed for discriminative problems
and generative tasks \cite{koller2009probabilistic}.
However, large-scale graphical models incur significant 
computational costs for both inference and learning.
Notable ways to overcome this difficulty are statistical sampling and
variational methods.
However, such methods typically require iterative computations,
and
the number of iterations to obtain reasonably accurate results is not known in advance.
In this paper,
we propose a learning formula for a limited class of the probabilistic graphical models, i.e. Ising models.
The proposed formula enables us
to determine the optimal model parameters for the approximate objective function
without iterations.

Ising models, which are also known as Boltzmann machines,
have binary random variables that interact via pairwise interactions.
Ising models have been widely used in statistical physics 
\cite{mezard1987spin},
and have been applied to many tasks that involve real-world data.
For example,
Ising models have been used to extract the latent combinatory control network
from observed correlations in the neural spike train data \cite{Schneidman2006},
protein structures \cite{Weigt2009},
and genome regulation \cite{Bailly-Bechet2010}.
In such applications, a model must learn an interaction matrix from the given data.
This problem has encouraged researchers to develop efficient methods
to solve the inverse statistical problems associated with Ising models \cite{Nguyen2017}.

Some non-iterative methods for inverse Ising problems have been developed.
For example, Bethe free energy, which is the objective function of a belief propagation algorithm,
has been analyzed in the Ising model \cite{Welling2003}.
Together with the linear response relation \cite{Kappen1998a},
an analytic expression of the correlation matrix has been obtained
as a function of the model parameter of the interaction matrix.
Surprisingly, it was later found that the analytic expression for the correlations
can be solved inversely for the interaction matrix \cite{Ricci-Tersenghi2011,Nguyen2012}.
This analytic inverse solution enables us to obtain
the optimal interaction matrix in Bethe approximation without iterations.

In this study, instead of the Bethe approximation,
we use the tree-reweighted (TRW) approximation to obtain
an approximate solution for the inverse Ising problem.
The TRW approximation was developed to improve the accuracy and convergence of
a belief propagation algorithm \cite{wainwright}.
Differing from the ordinary belief propagation algorithm,
TRW free energy is provably convex with respect to the variational parameters.
Moreover, the partition function computed using the TRW approximation 
gives a rigorous upper bound of the exact partition function.
Thus, TRW approximation gives a lower bound of the exact log-likelihood function
when learning a model \cite{Wainwright2003}.
Although the use of this lower bound for learning has been proposed
in the previous study,
it has not yet been applied to derive
an inverse formula in Ising model.

The reminder of this  paper is organized as follows.
In Section~\ref{s:ising}, 
we introduce the Ising model and its direct and inverse problems.
In Section~\ref{s:trw},
we briefly summarize the theorem and properties of TRW approximation.
Using TRW free energy, we analytically obtain the inverse formula
for the Ising model that optimizes the rigorous lower bound of the exact log-likelihood in Section~\ref{s:inv}.
In Section~\ref{s:rel}, we review related work and introduce
previously derived inverse formulae.
In Section~\ref{s:res},
we describe experiments conducted to compare the proposed inverse formula
and previous formulae.
Conclusions and suggestions for future work are provided in Section~\ref{s:con}.

 

\section{Ising model and Inverse Problem}\label{s:ising}
Ising models are undirected probabilistic graphical models
that describe pairwise interactions between binary spin variables
$s_i = \pm 1$.
Here, let $V$ be a set of binary variables
$\bs = \{ s_i \}_{i = 1}^{|V|}$ and
$E$ be a set of edges
$\left< ij \right> ( i,j \in V) $.
The energy function of an Ising model $G= (V, E)$ is defined by
\begin{align}
  E(\mathbf s) = - \sum_{\left< ij \right>\in E} J_{ij} s_i s_j -\sum_{i\in V}h_i s_i
  ,
\end{align}
where $J_{ij}$ and $h_i$ are parameters that represent
interaction strengths and biases (local external fields), respectively.
Given the energy function,
the probability distribution for the model is defined by the
Boltzmann distribution
\begin{align}
  \pr (\mathbf s) = \frac{1}{Z}\exp(-E(\mathbf s))
  =\exp\left( -E(\bs) - \Phi(\theta) \right)
  ,
  \label{e:pr}
\end{align}
where the partition function
\begin{align}
Z(\theta) = \sum_{\mathbf s} \exp(-E(\mathbf s))
,
\label{e:pf}
\end{align}
is defined as a function of the parameters $\theta = (J_{ij}, h_i)$,
and the log-partition function $\Phi(\theta )= \ln Z(\theta)$.

In practical applications of Ising models,
we are confronted by two difficult problems, i.e.,
the inference and learning problems.
The former is referred to as a direct problem,
and the latter is an inverse problem of the Ising model.

The direct problem is formulated as follows.
With fixed model parameter $\theta$,
one wants to compute expectation values, such as
$\left< s_i\right>$ and $\left< s_i s_j\right>$.
Computation of these expectation values
is generally intractable, because it
includes summations over numerous combinations of spin variables.
Note that this computational difficulty is
equivalent to that of the partition function:
If we know the function $\Phi(\theta)$ (or $Z(\theta)$),
we can readily obtain the expectation values.

In the inverse problem,
differing from the direct problem,
the goal is to infer the optimal parameter $\theta$
when a dataset $\bs^{(d)} (d=1,\dots, D)$ is given,
where the optimal parameter maximizes the log-likelihood function:
\begin{align}
  l(\theta)
  =
  \frac{1}{D}\sum_{d=1}^{D} \ln \Pr (\bs^{(d)} )
  =
  - \left< E \right>_D - \Phi(\theta)
  .
  \label{e:inv}
\end{align}
Here, $\left< \dots \right>_D$ denotes an expectation value with respect to
the given dataset
$\left< f \right>_D = \frac{1}{D}\sum_{d=1}^{D} f(\bs^{(d)})$.

Similar to the direct problem,
the partition function plays an important role in the inverse problem.
Because the log-likelihood function $l(\theta)$ is concave with respect to $\theta$ (Section~\ref{s:inv}),
the optimal parameters $\theta^*$ that maximize $l(\theta)$
can be obtained by solving $\del l/ \del \theta = 0$, i.e.,
\begin{align}
\left<s_i \right>_{D}
&=\frac{\partial \Phi}{\partial h_i}
,
\label{e:mm1}
\\
\left<s_i s_j \right>_{D}
&=\frac{\partial \Phi}{\partial J_{ij}}
.
\label{e:mm2}
\end{align}
These equations yield the well-known moment-matching conditions
for exact maximum likelihood estimates \cite{Amari1982}.

Note that this reformulation of the inverse problem
is an identity transformation of the problem,
and solving Eqs. (\ref{e:mm1}) and (\ref{e:mm2}) for parameter $\theta$ remains infeasible
because of the summation over all spin states in Eq.(\ref{e:pf}).

In the next section, we compute the moments in the model
using approximation.
Then, we solve the moment matching conditions for
the interaction parameter $J_{ij}$,
which yields the inverse formula as a function of the data statistics.

\section{Tree-reweighted Approximation}\label{s:trw}

The TRW approximation \cite{wainwright} provides
a systematic way to construct a rigorous upper bound
of the partition function.
Using this upper bound, approximate pseudo moments
that enable us to solve the moment-matching conditions (\ref{e:mm1}) and (\ref{e:mm2})  approximately
can be obtained.


Here, to introduce the TRW approximation,
we first define a spanning tree of a given graph and
a probability measure over a set of spanning trees.
Then,
we define TRW free energy and provide
the main theorem of TRW approximation.

\subsection{Spanning trees}\label{s:st}
A spanning tree $T \subset E$ of a given graphical model $G=(V, E)$
is defined by a tree graph in which
any two vertices in $V$ are connected via edges in $T$ and other vertices.
Let $\mathfrak T= \mathfrak T(G)$ be the set of all spanning trees of $G$.
Over the spanning trees in $\mathfrak T$,
we assign an arbitrary probability distribution $\rho(T)$
holding non-negativity and normalization, i.e., 
$\rho(T) \geq 0$ and $\sum_{T \in \mathfrak T} \rho(T) = 1$.

Next, we define a graphical model on each spanning tree $T \in \mathfrak T$.
Using the indicator function $\nu(T)$,
\begin{align}
  [\nu(T)]_{ij} =
  \begin{cases}
  1& \text{if $\left< ij \right> \in T$}\\
  0& \text{otherwise,}
  \end{cases}
\end{align}
we define the connectivity matrix of the model with spanning tree $T$
as $J(T) = \nu(T) J$.
By defining the parameter of the model with the spanning tree $T$ as
$\theta(T) = (J_{ij}(T), h_i)$,
we can write the log partition function with the spanning tree $T$
as $\Phi (\theta(T))$.

Finally, 
using the probability distribution $\rho (T)$,
we define the edge appearance probabilities $\rho_{ij}$
for an edge $\left< ij\right > \in E$ as follows:
\begin{align}
  \rho_{ij} = \sum_{T \in \mathfrak T} \rho(T) \nu_{ij}(T)
  .
\end{align}
Intuitively,
$\rho_{ij}$ represents a probability of the existence of the edge $\left< ij \right>$ in a tree $T$
if $T \in \mathfrak T$ is selected according to the probability $\rho(T)$.

\subsection{Tree-reweighted free energy}\label{s:fe}
Relative to the development of the TRW approximation,
we identify the  convexity of the log partition function
with respect to parameters $\theta$.
This is verified easily by computing the first and second derivatives,
for example, with respect to $J_{ij}$,
\begin{align}
  \frac{\del \Phi}{\del J_{ij}} &= \left< s_i s_j \right>
  ,
  \\
  \frac{\del^2 \Phi}{\del J_{ij} \del J_{kl}}
  &= \left< s_i s_j s_k s_l \right>
  -\left< s_i s_j \right>\left< s_k s_l \right>
  .
\end{align}
The second derivative is
the covariance matrix of $s_i s_j$ and $s_k s_l$.
Here, the covariance matrix is positive semidefinite,
which implies the convexity of $\Phi$.

Using the convexity of $\Phi$, the
upper bound of $\Phi$ can be obtained by applying Jensen's inequality as follows:
\begin{align}
  \Phi(\theta) &=
  \Phi\left( \sum_{T \in \mathfrak T} \rho(T) \theta(T) \right) \nonumber
  \\
  &\leq \sum_{T \in \mathfrak T} \rho(T) \Phi (\theta(T))
  \equiv \Phi^{\rm TRW}(\theta; \rho)
  ,
  \label{e:jensen}
\end{align}
where we used the fact that
$J_{ij}=\sum_{T\in \mathfrak T} \rho(T) J_{ij}(T)$.
A previous study \cite{wainwright} has demonstrated that
the upper bound $\Phi^{\rm TRW}(\theta; \rho)$ can be obtained
as a solution of the variational problem
that is defined by the TRW free energy.

In this variational problem,
variational parameters are the
pseudomarginals $q=\{q_i, q_{ij} \}$
that satisfy
\begin{align}
  q_i(s_i), q_{ij}(s_i, s_j) &\geq 0
  ,
  \\
  \sum_{s_i} q_i(s_i) &= 1,
  \\
  \sum_{s_i} q_{ij}(s_i, s_j) &= q_j(s_j)
\end{align}
for all $i \in V$ and $\left< ij \right> \in E$.

When a valid edge appearance probability $\rho_{ij}$ is given,
the TRW free energy is constructed by the energy and entropy terms as
\begin{align}
F^{\rm TRW}(q; \theta, \rho)
=
{\cal E}(q;\theta) - {\cal H}(q; \rho)
,
\label{e:trwfe}
\end{align}
where the energy term ${\cal E}$ and the entropy term ${\cal H}$ are defined as
\begin{align}
  {\cal E}(q;\theta)
  &=
-\sum_{<ij>} \sum_{s_i, s_j} s_i s_j J_{ij} q_{ij}(s_i, s_j)
- \sum_i \sum_{s_i} s_i h_i q_i(s_i)
,
\\
{\cal H}(q; \rho)
&=\sum_{\left< ij\right>} \rho_{ij}H_{ij}[q_{ij}]
+
 \sum_i (1-\sum_{j\in \left< ij \right>} \rho_{ij}) H_i[q_i]
,
\end{align}
respectively.
Here, we define the entropies of the pseudomarginals as
$H_{ij}[q]=-\sum_{s_i, s_j}q_{ij}(s_i, s_j)\ln q(s_i, s_j)$ and
$H_i[q]=-\sum_{s_i}q_{i}(s_i)\ln q(s_i)$.

The following theorem is proven \cite{wainwright} 
using the TRW free energy.
\begin{theorem}
  For a given $\rho$, the upper bound of Eq.(\ref{e:jensen})
  is obtained by the solution of the minimization problem:
  \begin{align}
    \Phi(\theta) \leq \Phi^{\rm TRW}(\theta;\rho) = -\min_q F^{\rm TRW}(q, \theta;\rho)
    .
  \end{align}
  Moreover, $F^{\rm TRW}$ is convex with respect to $q$: thus
  the optimal solution to this problem $q^*$ is unique.
  \label{t:trw}
\end{theorem}

\section{Approximate Solution to the Inverse Ising Problem}\label{s:inv}

\subsection{Rigorous lower bound of the objective}
In the TRW approximation, the exact objective function for learning (\ref{e:inv})
is approximated by
\begin{align}
  l^{\rm TRW}(\theta) = -\left< E \right>_D - \Phi^{\rm TRW}(\theta, \rho).
\end{align}
Note that,
with reference to Theorem\ref{t:trw},
this approximated objective is a rigorous lower bound for the
exact objective function:
\begin{align}
  l(\theta)\geq l^{\rm TRW} (\theta).
  \label{e:lb}
\end{align}
Consequently, maximizing $l^{\rm TRW}(\theta)$
results in maximization of the lower bound of
the exact objective function $l(\theta)$.

Since $\Phi(\theta)$ is a convex function,
$\Phi^{\rm TRW}(\theta, \rho)$ is also convex with respect to $\theta$.
This implies that $l^{\rm TRW} (\theta, \rho)$
is concave with respect to $\theta$
because the averaged energy term is linear in $\theta$.
Thus, the optimal solution is obtained by solving
$\del l^{\rm TRW}/\del \theta = 0$, which results in
the pseudo-moment matching conditions,
\begin{align}
\left<s_i \right>_D
&=\frac{\partial \Phi^{\rm TRW}}{\partial h_i}
,
\label{e:pmm1}
\\
\left<s_i s_j \right>_D
&=\frac{\partial \Phi^{\rm TRW}}{\partial J_{ij}}
.
\label{e:pmm2}
\end{align}
Note that these equations are approximations
of the exact moment matching conditions (\ref{e:mm1}) and (\ref{e:mm2}).

\subsection{Analytic Solution in the TRW Approximation}
Next, we give the explicit solution
of the pseudo-moment matching conditions(\ref{e:pmm1}) and (\ref{e:pmm2}).
Given that all random variables are binary,
we parameterize pseudomarginals $q_i, q_{ij}$
by mean value $m_i$ and covariance $c_{ij}$ as follows:
\begin{align}
  q_i(s_i) &= \frac{1}{2}(1+m_i s_i)
  ,\label{e:qi}
  \\
  q_{ij}(s_i, s_j) &= \frac{1}{4}
  \left(
  (1+m_i s_i) (1+m_j s_j) +c_{ij}s_i s_j
  \right)
  .
  \label{e:qij}
\end{align}  
By substituting these equations into the
definition of TRW free energy (\ref{e:trwfe}),
the optimal solutions $m^*_i, c^*_{ij}$ are obtained
by solving
$\del F^{\rm TRW}/\del m_i = \del F^{\rm TRW}/\del c_{ij} = 0$.

By applying cavity methods \cite{Mezard2001}\cite{Mezard2003},
we derive a self-consistency equation for $m_i$
by transforming the equations $\del F^{\rm TRW}/\del m_i = 0$ and $\del F^{\rm TRW}/\del c_{ij} = 0$.
Following the derivation in Bethe approximation \cite{Welling2003} \cite{Ricci-Tersenghi2011},
we obtain the self-consistency equation of TRW approximation as:
\begin{align}
m_i
=
\tanh \left[
h_i + \sum_j \rho_{ij} \atanh \left( \tilde t_{ij} f(m_j, m_i, \tilde t_{ij} )\right)
\right]
,
\label{e:sce}
\end{align}
where $\tilde t_{ij}=\tanh J_{ij}/\rho_{ij}$ and
\begin{align}
  &f(m_1, m_2, t)
    \nonumber
  \\
=&
\frac{1-t^2 -\sqrt{(1-t^2)^2-4t(m_1-m_2t)(m_2-m_1t)}}
     {2t(m_2-m_1t)}
     .
\end{align}

Now, consider the pseudo-moment matching conditions.
Eq.~(\ref{e:pmm1}) simply states that the mean value of the model
should be matched to that computed by data, i.e.,
$m^*_i = \left< s_i \right>_D$.
We rewrite $\left< s_i \right>_D = \hat m_i$ for simplicity.
Note that this should not be confused with $m^*_i$.

Eq.~(\ref{e:pmm2}) requires more careful treatment.
Using the linear response relation \cite{Kappen1998a},
we find the following relation between derivatives,
\begin{align}
  \frac{\del^2 \Phi^{\rm TRW}}{\del h_j \del h_i}
  =
  \frac{\del \Phi^{\rm TRW}}{\del J_{ij}}
  -
  \frac{\del \Phi^{\rm TRW}}{\del h_i}\frac{\del \Phi^{\rm TRW}}{\del h_j}
  .
\end{align}
Together with the pseudo-moment matching conditions (\ref{e:pmm1}) and (\ref{e:pmm2}),
and the fact that $m^* = \del \Phi^{\rm TRW}/ \del h_i$,
we obtain:
\begin{align}
  \frac{\del m^*_i}{\del h_j}
  =
  \left< s_i s_j \right>_D - \left< s_i \right>_D \left< s_j \right>_D = C_{ij}
  ,
  \label{e:lr}
\end{align}
where we have defined the covariance of data $C_{ij} = \left< s_i s_j \right>_D - \left< s_i \right>_D \left< s_j \right>_D$.
This equation states that, to include the information about the covariance of the data,
we must know the $h_j$ derivative of the fixed-point solution $m^*_i$.

Taking the $m_j$ derivative to the both sides of Eq.(\ref{e:sce}),
we obtain the following.
\begin{align}
  \delta_{ij}
  =
  \left( 1- m_i^2 \right)
  \left[ \frac{\del h_i}{\del m_j}
    + \rho_{ij}
    \frac{\tilde t_{ij} \frac{\del}{\del m_j} f(m_j, m_i, \tilde t_{ij}) }
         {1-(\tilde t_{ij} f(m_j, m_i, \tilde t_{ij}))^2}
         \right]
  .
  \label{e:sce_del}
\end{align}
Since the linear response relation (\ref{e:lr}) states that:
\begin{align}
  \frac{\del h_i}{\del m_j} = [C^{-1}]_{ij},
\end{align}
we obtain the following from Eq.(\ref{e:sce_del}), for any $i\neq j$, 
\begin{align}
  0
  =
  [C^{-1}]_{ij}
    + \rho_{ij}
    \frac{\tilde t_{ij} \frac{\del}{\del m_j} f(m_j, m_i, \tilde t_{ij}) }
         {1-(\tilde t_{ij} f(m_j, m_i, \tilde t_{ij}))^2}
 .
\end{align}
Finally, replacing $m_i \to \hat m_i$ and solving this equation for $J_{ij}$,
we obtain an approximate solution to the inverse Ising problem:
\begin{widetext}
\begin{align}
  J^{\rm TRW}_{ij}
  =
-\rho_{ij}
 \atanh \left[
\frac{1}{2(\tilde C^{-1})_{ij}}
\sqrt{\tilde D_{ij}}
+
\frac{1}{2(\tilde C^{-1})_{ij}}
\sqrt{( \sqrt{\tilde D_{ij}} -2\hat m_i \hat m_j (\tilde C^{-1})_{ij} )^2 - 4(\tilde C^{-1})_{ij}^2}
-\hat m_i \hat m_j
\right]
\label{e:inv_trw}
\end{align}
\end{widetext}
where, 
$(\tilde C^{-1})_{ij}=(C^{-1})_{ij}/\rho_{ij}$
and 
\begin{align}
\tilde D_{ij}
=1+4(1-\hat m_i^2)(1-\hat m_j^2)(\tilde C^{-1})^2_{ij}
.
\end{align}

\section{Related Work}\label{s:rel}

\subsection{Inference and learning in TRW approximation}

The TRW approximation was introduced \cite{wainwright}
to improve belief propagation algorithm for general graphs with loops.
This belief propagation algorithm,
which was first used for the exact inference method for
the models without loops \cite{pearl2014probabilistic},
was later applied to approximate inference of models with loops \cite{Murphy1999}.
Although there are many examples where belief propagation gives
reasonable results,
it has been found that the belief propagation algorithm
cannot converge in some cases.
Theoretically,
it has been pointed out that Bethe free energy,
which is the objective function of belief propagation \cite{Yedidia2000a},
is not always convex for general models \cite{Heskes2003},
which explains the divergence and cyclic behavior of
the belief propagation iteration.
On the other hand,
the TRW free energy is 
convex with respect to the variational parameters.
This fact improves the convergence of the iteration.
Another advantage is that the TRW approximation
can give the rigorous upper bound of the exact partition function.

An early attempt to use TRW approximation for learning graphical models
can be found in the literature \cite{Wainwright2003}.
The authors of \cite{Wainwright2003} pointed out
that TRW approximation gives a rigorous lower bound for
the exact objective function (Eq.(\ref{e:lb})),
and that maximization of the TRW objective function can be achieved
by using the pseudo-moment matching conditions
(Eqs.(\ref{e:pmm1}) and (\ref{e:pmm2})).
In contrast to our approach,
they used a variant of the belief propagation algorithm
to compute the pseudo moments.
Although this approach is widely applicable to various models
other than Ising models,
iterative computation is required until the belief propagation algorithm converges.
Furthermore, convergence of the belief propagation algorithm is not guaranteed
in general.
In this paper, we avoid this difficulty by directly calculating
the analytic expression for the pseudo moments
that minimize the TRW free energy.

\subsection{Existing inverse formulae}
Various inverse formulae for the inverse Ising problem
have been developed, and
some of them are presented in this subsection
because they are compared to the proposed inverse formula in the next
section.

In the independent-pair (IP) approximation \cite{Roudi2009a},
only effects of a pair of spins $s_i$ and $s_j$ are taken into account
in computation of the interaction $J_{ij}$,
which leads to the formula:
\begin{align}
  &J_{ij}^{\rm IP}
  \nonumber
  \\
  =&
  \ln 
  \frac{((1+\hat m_i)(1+\hat m_j)+c_{ij})((1-\hat m_i)(1-\hat m_j)+c_{ij})}
  {((1+\hat m_i)(1-\hat m_j)-c_{ij})((1-\hat m_i)(1+\hat m_j)-c_{ij})}
  \label{e:inv_ip}
  .
\end{align}
Note that the IP approximation corresponds to the Bethe approximation
without using the linear response relation.

Another approach is to use the series of the mean-field approximations \cite{Kappen1998a}\cite{Tanaka1998}.
The most advanced approximation used to derive the inverse formula
is Bethe approximation \cite{Ricci-Tersenghi2011}\cite{Nguyen2012}.
The result is given by setting $\rho_{ij} = 1$ in Eq.~(\ref{e:inv_trw}),
\begin{align}
  J^{\rm BA}_{ij} =
  \left. J^{\rm TRW}_{ij} \right|_{\rho_{ij}=1}
  ,
  \label{e:inv_bethe}
\end{align}
because the TRW free energy is reduced to Bethe free energy
by setting $\rho_{ij}=1$.
Note that this does not mean that the TRW approximation includes
the Bethe approximation:
The choice of $\rho_{ij} = 1$ is invalid unless the graph is a tree.

In \cite{Sessak2009}, another inverse formula is developed 
using small-correlation expansion of the entropy function.
After resumming over the relevant diagrams,
the Sessak-Monasson (SM) approximation is obtained as follows:
\begin{align}
  J^{\rm SM}_{ij}= - (C^{-1})_{ij} + J^{\rm IP}_{ij}
  -\frac{C_{ij}}{(1-\hat m_i^2)(1-\hat m_j^2)-(C_{ij})^2}
  \label{e:inv_sm}
  .
\end{align}

\section{Numerical Results}\label{s:res}
In this section, we show the results of the numerical experiments
by comparing the TRW inverse formula, Eq.(\ref{e:inv_trw}), and
the other inverse formulae, Eqs.(\ref{e:inv_ip}), (\ref{e:inv_bethe}) and (\ref{e:inv_sm}).

In subsection~\ref{s:recons}, using the Ising models with the artificial parameters,
we reconstruct the parameters by the inverse formulae.
We compare the accuracy of the methods by measuring the reconstruction errors.

In subsection~\ref{s:spike}, using the real-world neural spike train data,
we estimate effective synaptic connections between the recorded neurons
as an interaction matrix of an Ising model \cite{Schneidman2006}.
Because we do not know the exact interactions matrices,
we evaluate the accuracy of the inverse formulae 
by comparing the inferred interaction matrices with that computed
by the gradient ascent algorithm.

\subsection{Artificial data}\label{s:recons}

To evaluate the accuracy of the proposed formula,
we attempted to reconstruct an interaction matrix from
the statistics of sampled data.
The experimental method  is described as follows:
We built Ising models with fixed graph structures and parameters.
Using the Monte Carlo sampling method,
we computed the mean value
$\hat m_i = \left< s_i \right>_D $ and
covariance
$C_{ij} = \left< s_i s_j \right>_D -\hat m_i \hat m_j$ in the models.
Finally, by substituting $m_i$ and $c_{ij}$
to the inverse formulae,
we reconstructed the interaction matrices $J_{ij}$.
For the Monte Carlo simulation,
we used the PyMC3 package \cite{salvatier2016probabilistic}.

Here, we used three types of graph structures, i.e.,
two-dimensional grid, three-dimensional grid,
and fully-connected graphs.
The numbers of random variables were set as
$7\times 7$, $4\times 4\times 4$, and 16
for the two-dimensional grid,
three-dimensional grid, and
fully-connected graph, respectively.
For each graph,
we set two types of interaction matrices for attractive and mixed interactions.
For both interactions, 
we used uniform distribution $u$ with given interaction strength $\omega >0 $
for all pairs $\left< ij \right> \in E$ as
\begin{align}
  J_{ij} \sim
  \begin{cases}
    u[0, \omega]& \text{(attractive)}\\
    u[-\omega, \omega]& \text{(mixed)}.
  \end{cases}
\end{align}
The bias parameter $h_i$ was set by
$h_i \sim u[-0.05, 0.05]$ for both attractive and mixed settings.

For the inverse formulae,
we compared the IP approximation (\ref{e:inv_ip}),
Bethe approximation (\ref{e:inv_bethe}),
SM approximation (\ref{e:inv_sm}),
and TRW approximation (\ref{e:inv_trw}).
For TRW approximation,
we set $\rho_{ij}=(|V|-1)/|E|$ \cite{wainwright},
which corresponds to assigning
the probability to the edges uniformly
(see the discussion in the conclusion).
To measure the error in the reconstructed interaction,
we compared it to the true interaction $J_{ij}^{\rm true}$
using the normalized distance:
\begin{align}
\Delta_J=
\sqrt{
\frac{\sum_{<ij>}(J_{ij} - J^{\rm true}_{ij})^2}
{\sum_{<ij>} (J^{\rm true }_{ij})^2}
}
.
\label{e:delta}
\end{align}
The smaller $\Delta_J$ becomes, 
the better the reconstruction is.

\begin{figure*}[!t]
\centering
\subfloat[]{\includegraphics[width=2.5in]{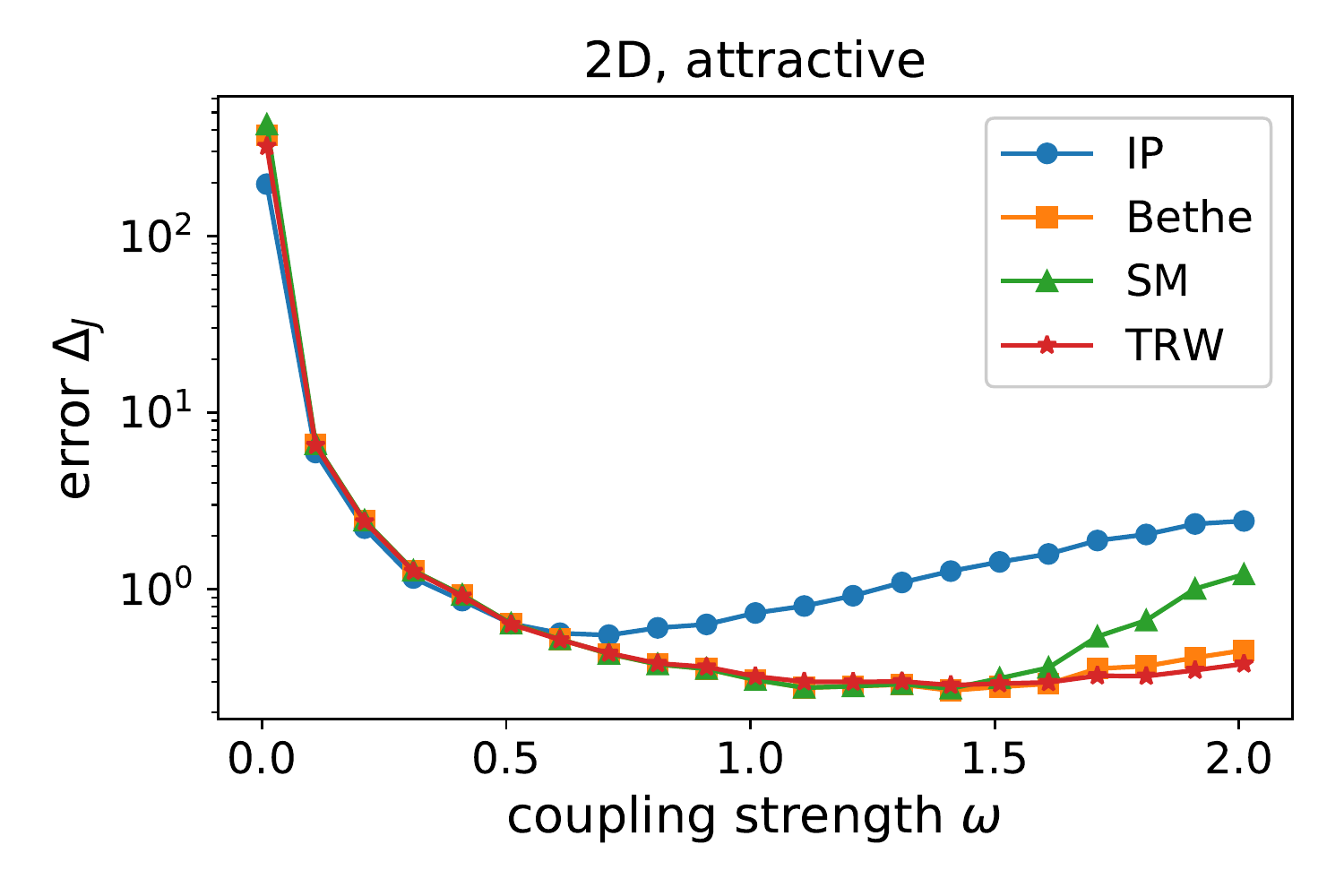}%
}
\hfil
\subfloat[]{\includegraphics[width=2.5in]{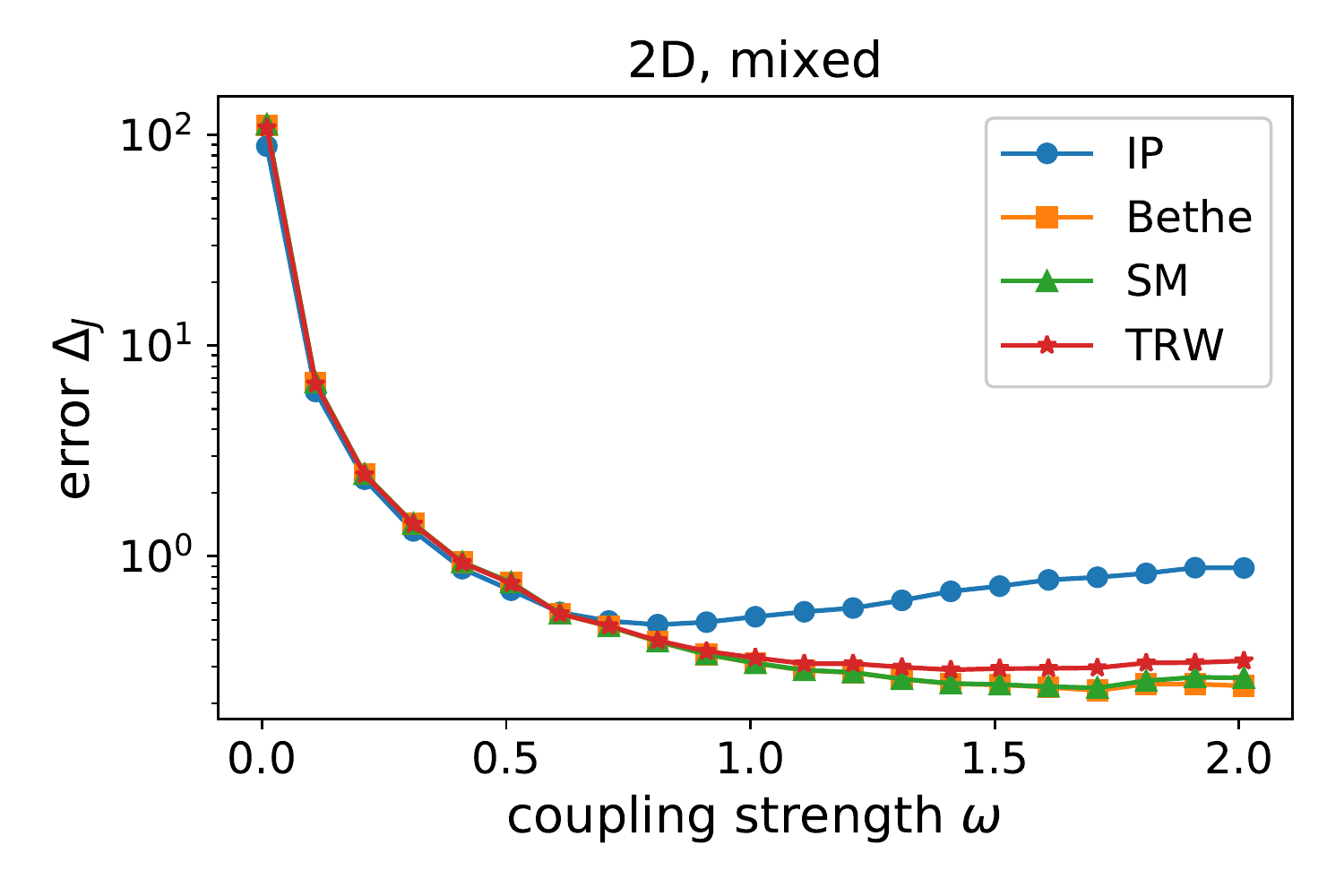}%
}
\\
\subfloat[]{\includegraphics[width=2.5in]{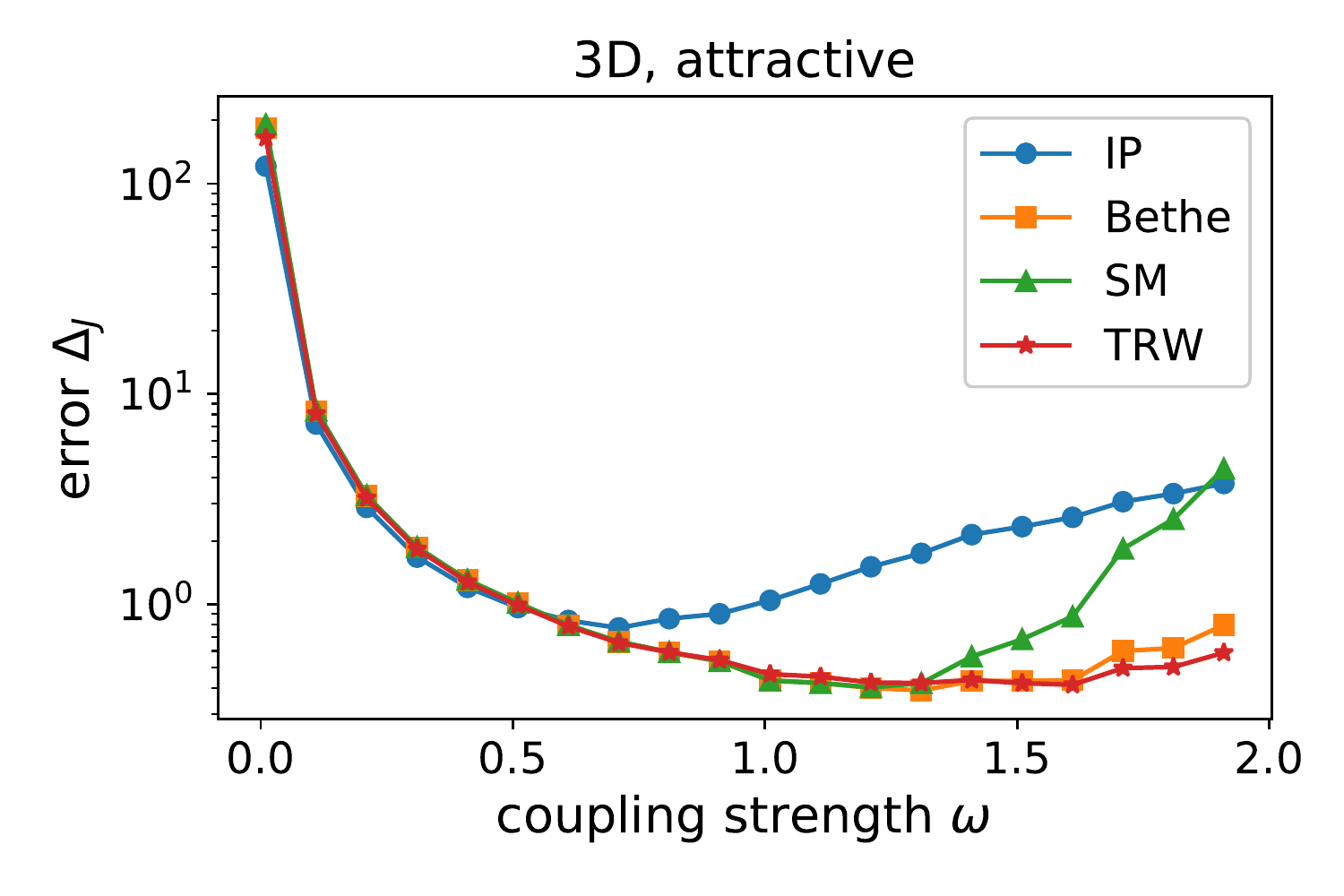}%
}
\hfil
\subfloat[]{\includegraphics[width=2.5in]{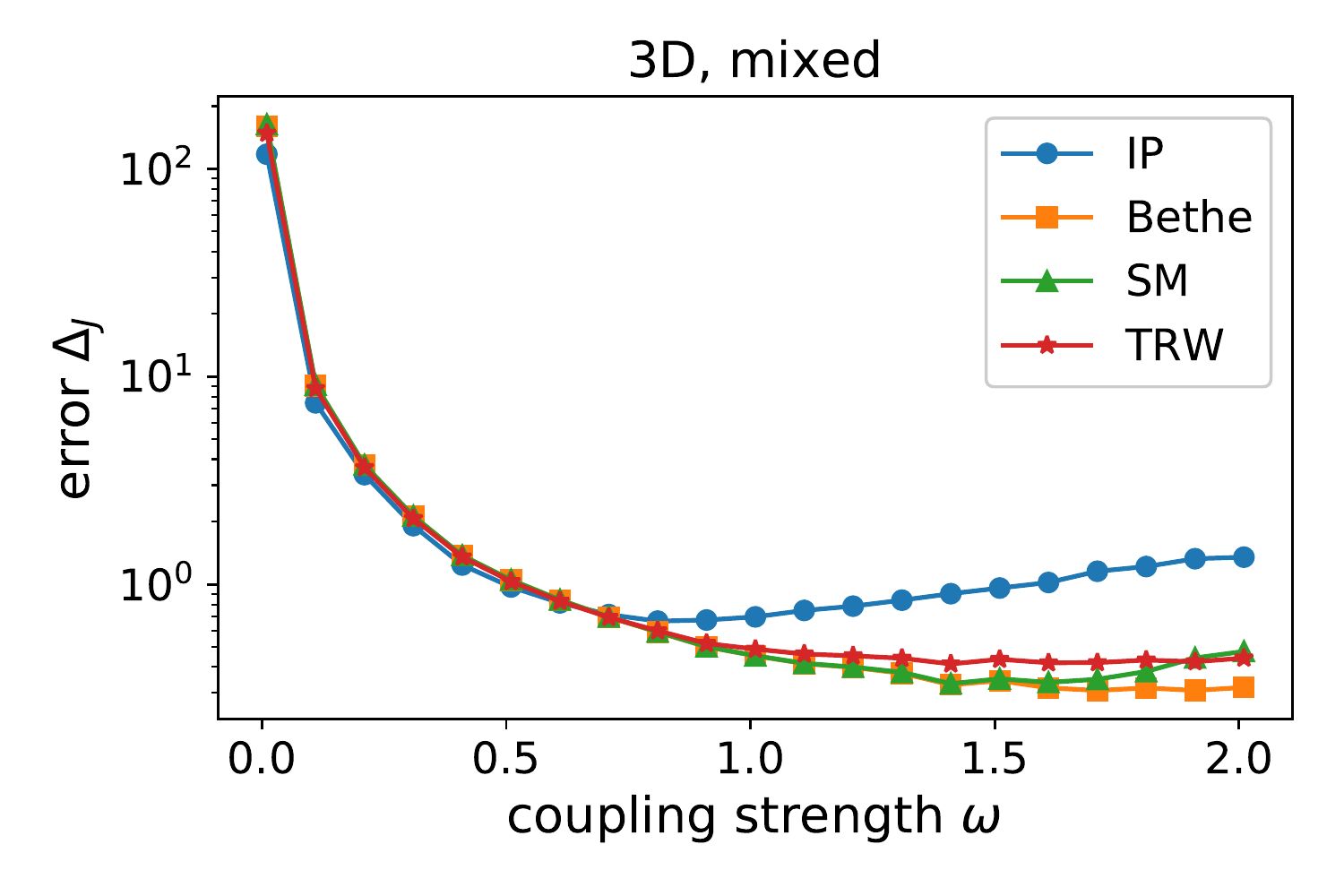}%
}
\\
\subfloat[]{\includegraphics[width=2.5in]{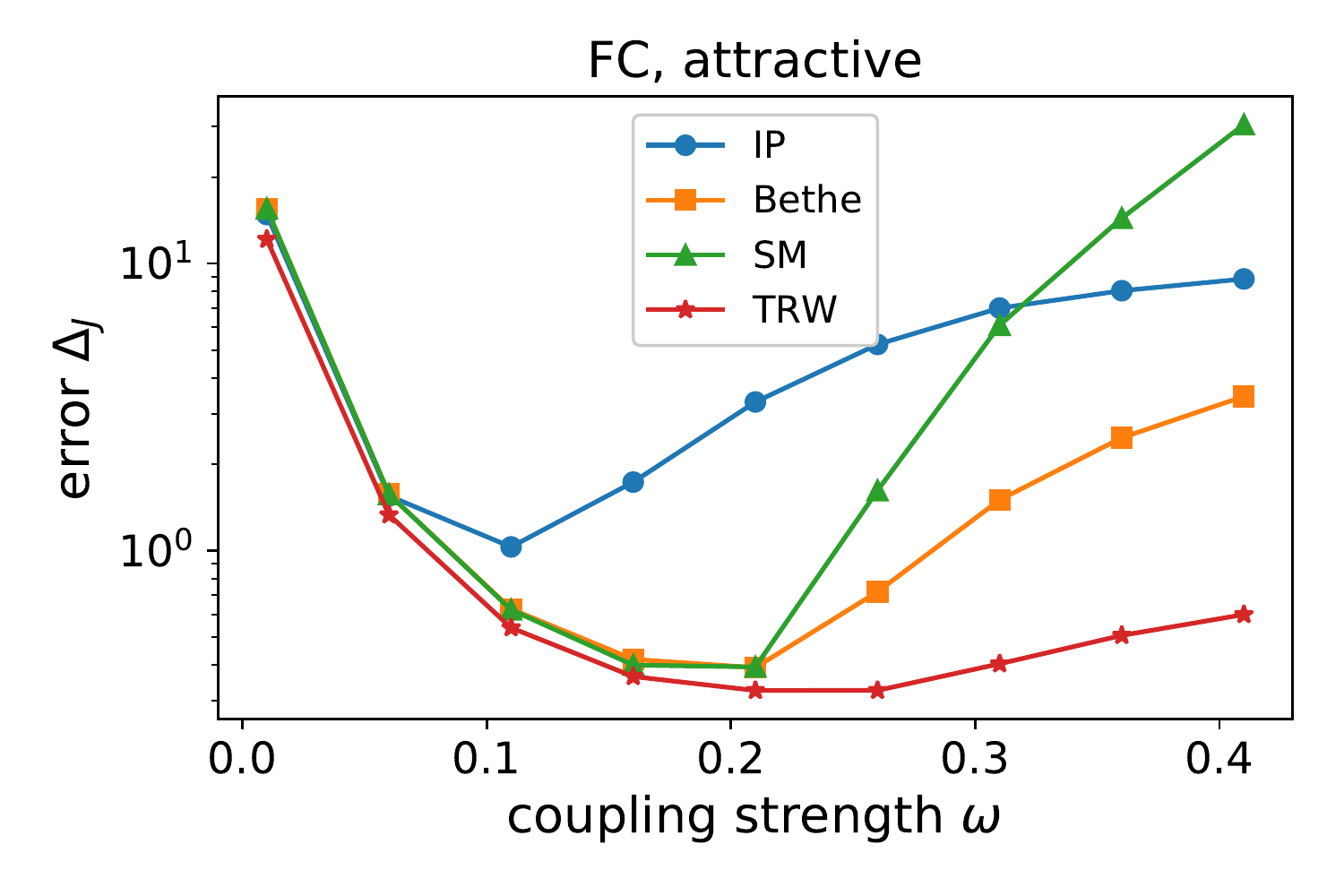}%
}
\hfil
\subfloat[]{\includegraphics[width=2.5in]{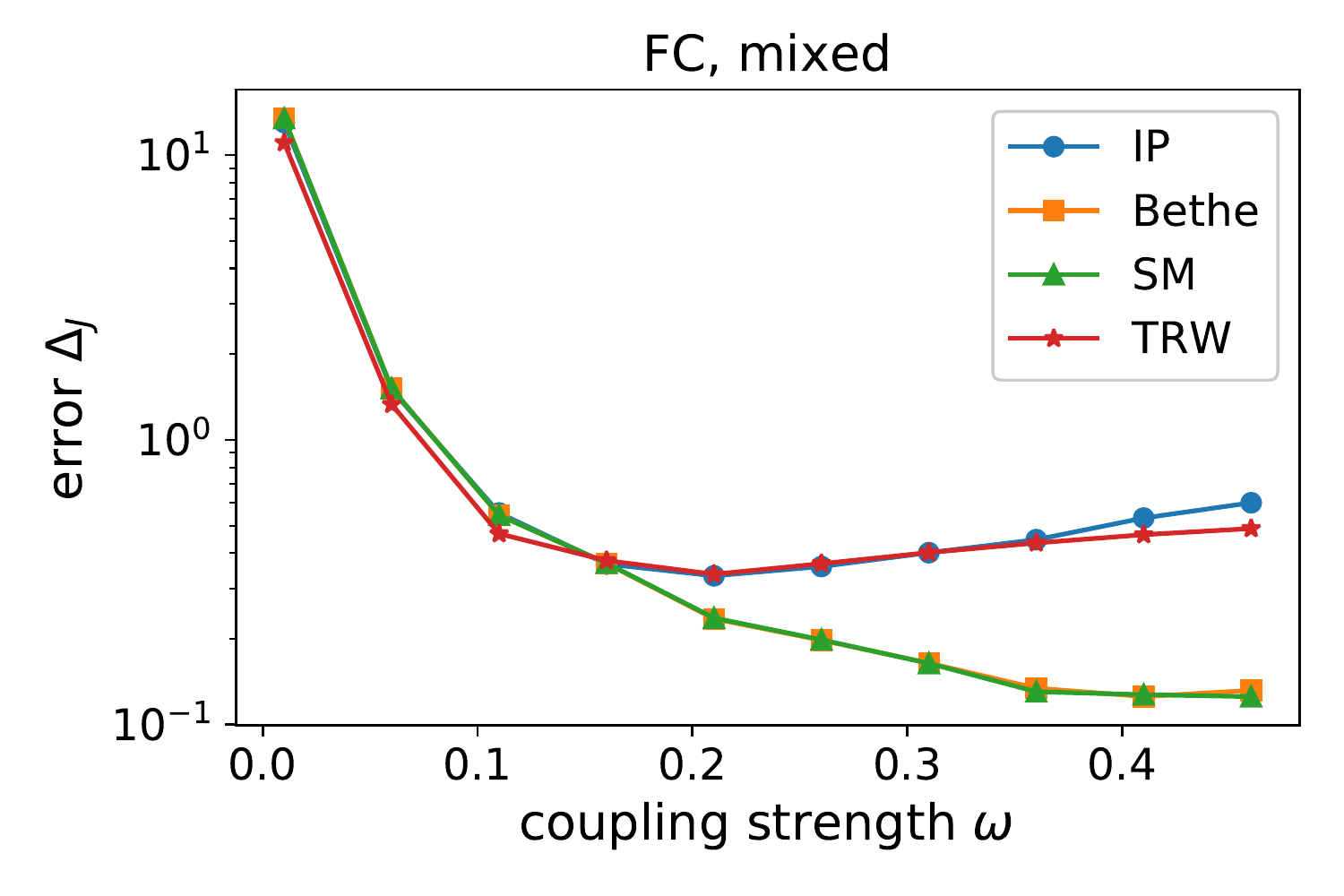}%
}
\caption{%
  Reconstruction errors with coupling strengths $\omega$
  for (a) and (b) two-dimensional grid,
  (c) and (d) three dimensional grid, 
  and (e) and (f) fully-connected graphs.
  The interaction matrices are attractive for (a), (c) and (e),
  and mixed for (b), (d), and (f).
}
\label{f:recons}
\end{figure*}

The reconstruction results for
the two-dimensional grid,
three-dimensional grid,
and fully-connected graph are shown in Fig.~\ref{f:recons}.
For all graph structures,
the error measurement $\Delta_J$ becomes large in
small interaction strength $\omega$ regions
irrespective of the inverse formulae
because, in such small $\omega$ regions,
the statistical uncertainty of the mean and covariance
primarily dominates the errors \cite{Ricci-Tersenghi2011}\cite{Marinari2010}.

When $\omega$ is apart from zero,
the differences between the reconstruction errors become large.
For all settings, the IP approximation typically gave the worst results.
For the attractive interactions,
the proposed formula based on TRW approximation gave the best accuracy.
For the fully-connected graph,
in which reconstruction is the most severe (note the scales in Fig.~\ref{f:recons}),
the proposed formula demonstrated reasonable errors around $\omega \sim 4$.

However, for the mixed interactions,
Bethe approximation gave the best results.
Here, TRW approximation was comparable to the Bethe and
SM approximations for grid graphs,
and TRW approximation was
better than SM approximation in the three-dimensional graph
around $\omega \sim 2$.
For the fully-connected model,
TRW approximation gave a worse result,
which is close to that of IP approximation.

\subsection{Real-world neural data}\label{s:spike}

In the next experiment,
we evaluated the inverse formulae with a task
to infer latent effective networks of neurons \cite{Schneidman2006},
using real-world neural spike train data.
Neural spike train data are composed
by multidimensional series of spiking times.
These data are converted to a set of spin states,
to witch we attempted to fit the Ising model (\ref{e:pr})
by optimizing parameters $J_{ij}$ and $h_i$.
The inferred interaction matrix $J_{ij}$ represents
effective synaptic connectivity between neurons $i$ and $j$,
i.e., if $J_{ij}$ is positive, the connection between them is excitatory,
and if $J_{ij}$ is negative, the connection is inhibitory.
The absolute value of $J_{ij}$ represents strength of the connection.

In this task, we used neural spike train data \cite{Lefebvre2008},
from CRCNS.org (ret-1) \cite{crcns}.
The data were collected by multi-electrode array recordings
of the retinal ganglion cells of mice in various conditions.
From the whole data,
we selected three recordings 
(in \texttt{20080516\_R1.mat}),
because in these recordings, the number of recorded neurons, seven,
was large enough not to give a trivial inverse problem, and
not too large for applying the inverse formulae to obtain reasonable results
(see the number of spins in the experiment in Sec.~VI.~A).

The method of this experiment was as follows \cite{Schneidman2006}\cite{Roudi2009a}:
First, we converted the spike train data to a set of spin states.
By giving bin size $\tau$,
we assigned a spin state $s_i^{(t)}$
to the $i$-th neuron in the $t$-th bin.
Here, if there were one or more spikes, we set $s_i^{(t)}=+1$,
otherwise $s_i^{(t)}=-1$.
We used a bin size $\tau = 1$ms.
Next, using the obtained spin states, we computed
the statistics
$\hat m_i = \frac{1}{T} \sum_{t=1}^{T} s_i^{(t)}$
and
$C_{ij} = \frac{1}{T} \sum_{t=1}^{T} s_i^{(t)} s_j^{(t)} - \hat m_i \hat m_j$.
Finally, with these statistics, we inferred the interaction matrix $J_{ij}$
of a fully-connected model using the inverse formulae.

For this experiment, 
we compared the IP approximation,
Bethe approximation,
SM approximation,
and TRW approximation for the inverse formulae.
For TRW approximation, we set $\rho_{ij}=(|V|-1)/|E|$.

To evaluate the estimates of the inverse formulae,
we compared the inferred interaction matrices
to what was obtained by the gradient ascent algorithm,
because we did not know the exact interaction matrix,
unlike the artificial models in the previous experiment.
In each update of the gradient ascent algorithm,
we computed the gradients $\del l/\del h_i$ and $\del l/\del J_{ij}$
by running the Monte Carlo algorithm in one hundred steps using the PyMC3 package.
We updated the parameters ten thousand times with the learning rate $\alpha = 0.1$ \cite{Roudi2009a}.
It took about ninety minutes to compute one interaction matrix.

We measured the distances between the interaction matrices
computed by the inverse formulae and by the gradient ascent algorithm
measured by Eq.(\ref{e:delta}) with $J^{\rm true}$
replaced by the gradient ascent result.

The distances $\Delta_J$
between the estimates obtained by the inverse formulae
and the gradient ascent algorithm measured by Eq.~(\ref{e:delta})
are shown in Table~\ref{t:ret}.
As can be seen, for all three datasets,
the developed TRW approximation gave the smallest
distance from the gradient ascent results.
Somewhat surprisingly,
SM approximation always gives the worst result.
The hierarchy of the accuracy of the approximations
was consistent with the previous reconstruction experiments
in the fully-connected graph with the attractive interaction
( Fig.~\ref{f:recons} (e)).
Let us remark that it took less than 10 ms
to compute an interaction matrix using the inverse formulae.

\begin{table}[!t]
\renewcommand{\arraystretch}{1.3}
\caption{Distances of inferred interaction matrices from gradient ascent results}
\label{t:ret}
\centering
\begin{tabular}{l||c|c|c}
{\bf Sample number} & 1 & 2 & 3\\
\hline
\hline
$\Delta_{J}$ & & &\\
IP & 0.91 & 0.88 & 0.72\\
\hline
Bethe & 0.57 & 0.58 & 0.60\\
\hline
SM & 1.23 & 1.17 & 1.19\\
\hline
TRW & 0.40 & 0.39 & 0.49\\
\end{tabular}
\end{table}

\begin{figure*}[]
\centering
\subfloat[IP]{\includegraphics[width=2.5in]{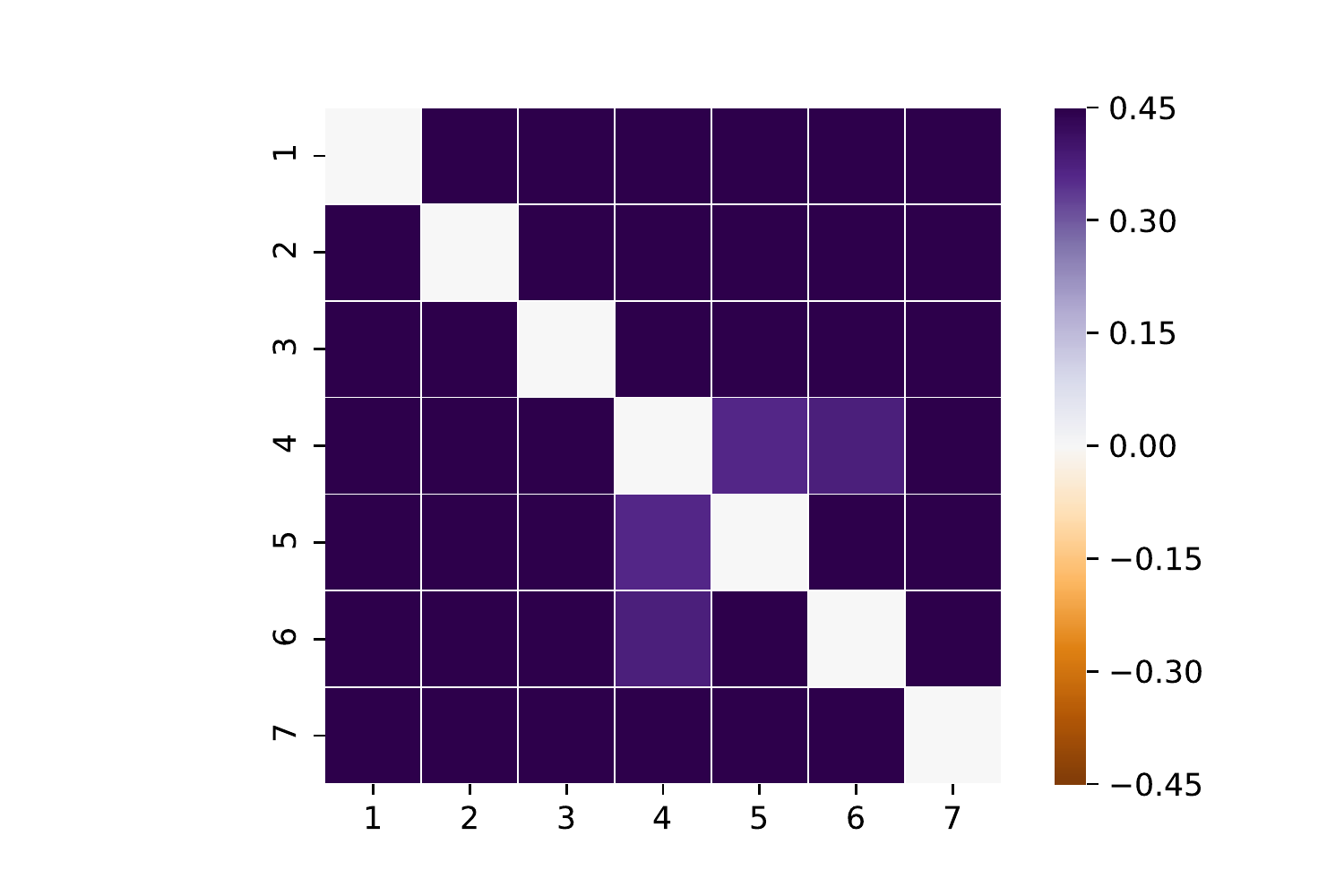}}
\hfil
\subfloat[Bethe]{\includegraphics[width=2.5in]{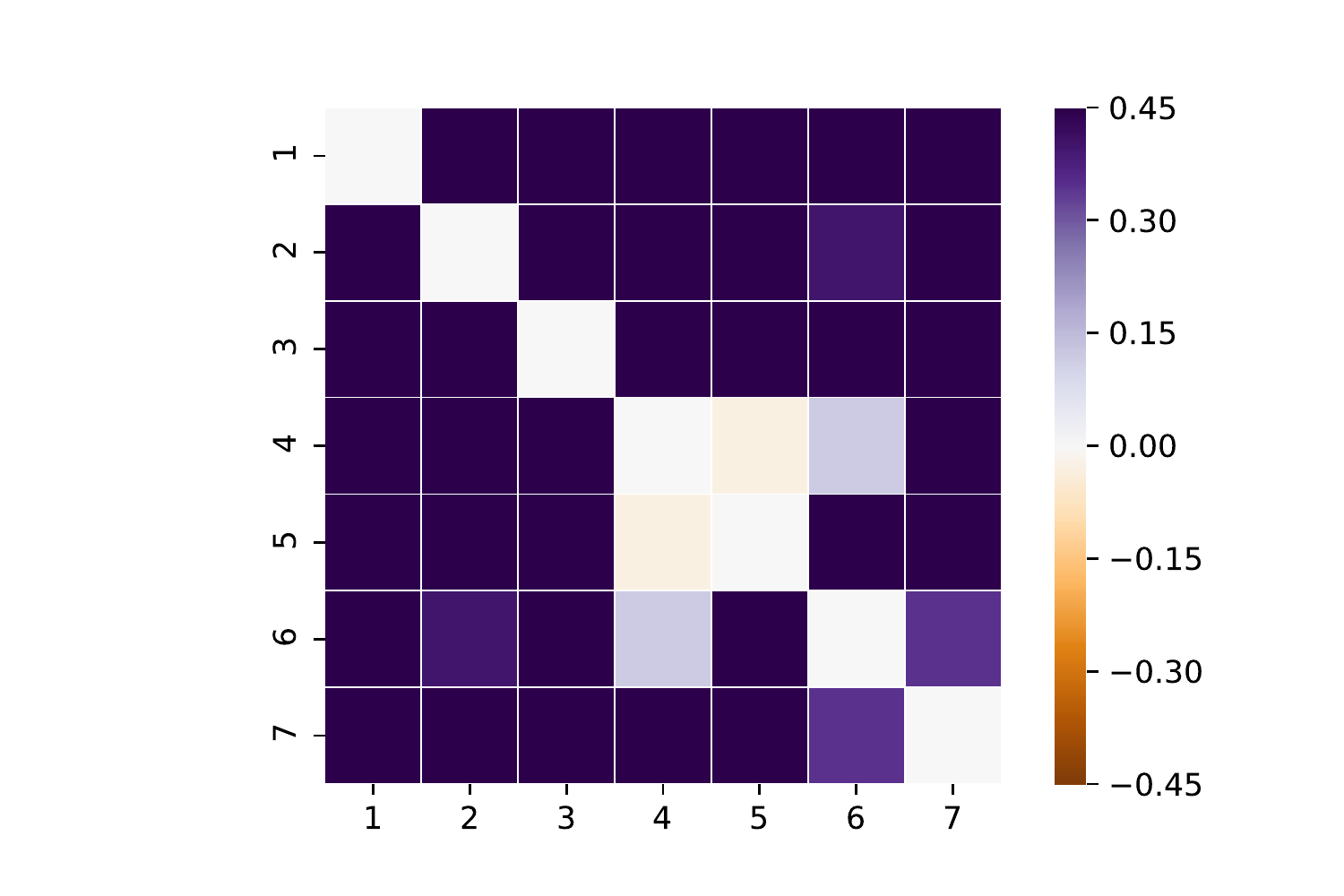}}
\\
\subfloat[TRW]{\includegraphics[width=2.5in]{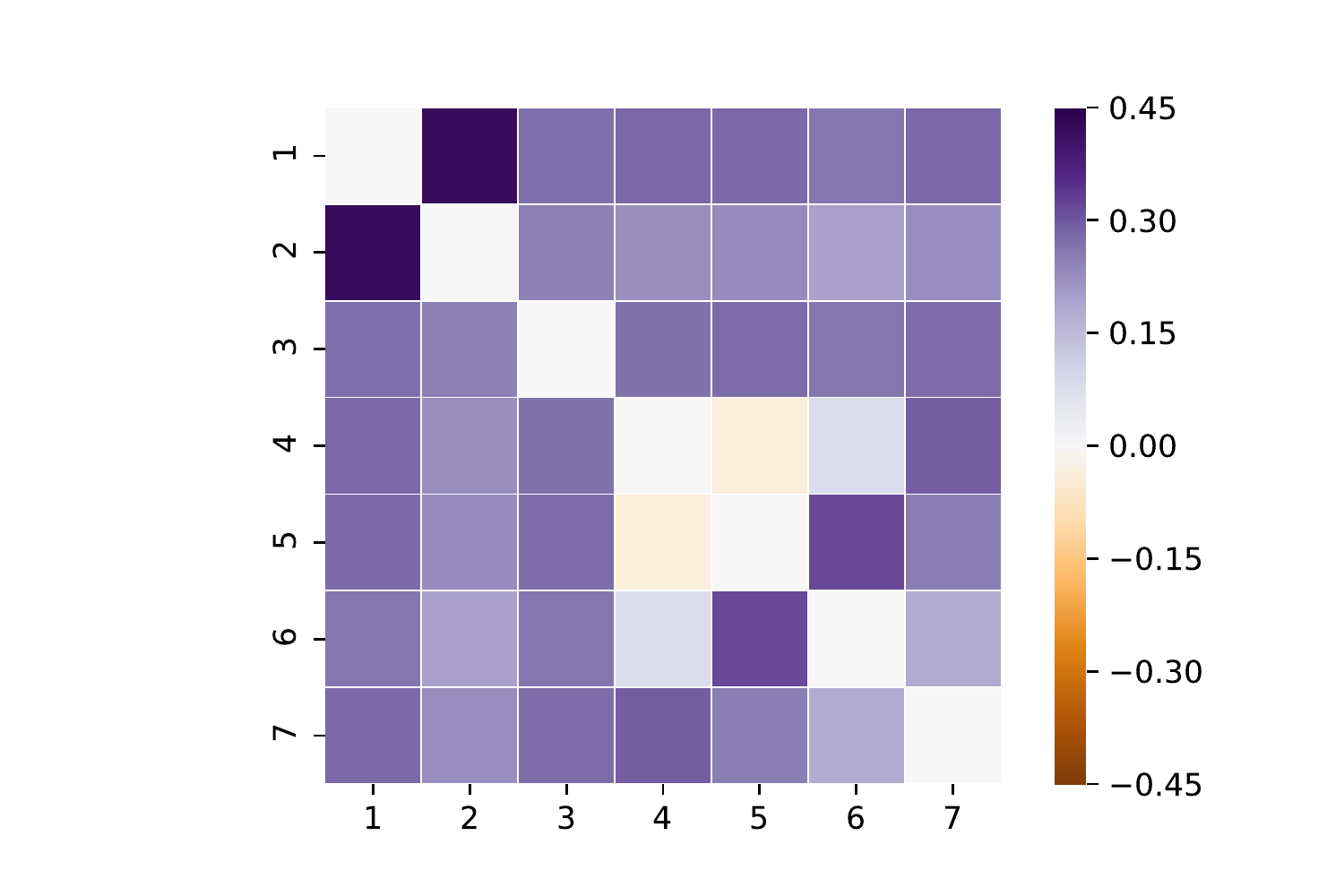}}
\hfil
\subfloat[gradient ascent]{\includegraphics[width=2.5in]{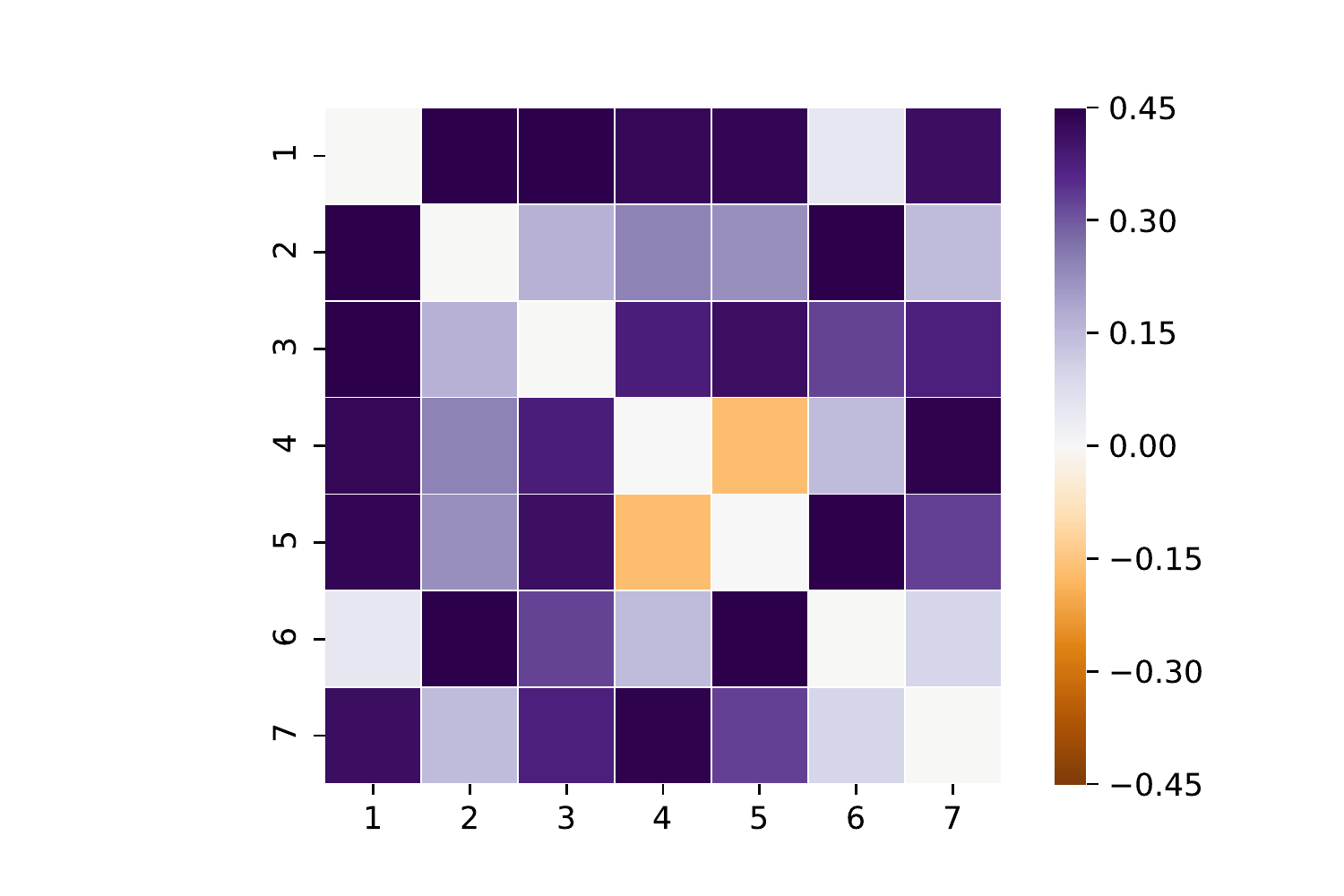}}
\caption{%
  Learned interaction matrices $J_{ij}$
  for the neural spike train data obtained by
  (a) IP approximation,
  (b) Bethe approximation,
  and (c) TRW approximation.
  The result obtained by the gradient ascent algorithm is shown in (d) for comparison.
}
\label{f:ret}
\end{figure*}

Examples of the resulting interaction matrices
of the IP, Bethe, and TRW approximations and of the gradient ascent algorithm are visualized in Fig.~\ref{f:ret}.
Note that most elements of the matrix are positive
in the gradient ascent result.
Therefore, the problem can be considered similar to the attractive case in
the fully-connected graph (Fig.~\ref{f:recons}(e)),
as suggested by the hierarchy of $\Delta_J$'s in Table~\ref{t:ret}.
Compared to the gradient ascent result,
the IP and Bethe approximations tended to overestimate
the interactions.
Furthermore, the IP approximation could not
reproduce the negative element in the interaction
between the fourth and fifth neurons.
While both Bethe and TRW approximations could reproduce
the negative elements, TRW approximation gave
a result that was the most similar to that of the
gradient ascent as measured by Eq.~(\ref{e:delta}).

\section{Conclusion}\label{s:con}
Aiming at faster and more accurate learning,
we have developed a new, iteration-free formula for the inverse Ising problem.
Following a previous study \cite{Ricci-Tersenghi2011},
we combined the linear response relation and the pseudo-moment matching
conditions to derive the inverse formula.
A remarkable difference from that is
that we used tree-reweighted free energy rather than Bethe free energy.
An advantage of using tree-reweighted free energy is that
we can optimize the rigorous lower bound of the exact objective function.
We analytically obtained the resulting inverse formula (\ref{e:inv_trw}),
which gives the interaction matrix as the function of 
the edge appearance probability $\rho_{ij}$
as well as the statistics of the input dataset $\hat m_i$ and $C_{ij}$.
Using this formula,
we can compute the approximate interaction matrix in the same computational complexity as
in Bethe approximation (\ref{e:inv_bethe}).

We compared the proposed inverse formula to other
formulae in interaction reconstruction experiments
with various graph structures and interaction matrices.
We found that the proposed formula gave the best accuracy
in models with attractive interaction matrices
(i.e. the elements of the interaction matrices were positive).
In particular, for fully-connected graphs,
the proposed formula showed overwhelmingly good reconstructions
compared to Bethe and Sessak-Monnason approximations.
In contrast, for mixed interaction matrices(i.e. the elements were both positive and negative),
the best approximation was obtained by Bethe approximation in most cases:
however, the proposed formula gave comparable results in models with grid graphs.

We also applied our formula to real-world neural spike train data.
In a task to infer latent effective networks,
our formula gave a result most similar to
that obtained by the gradient ascent algorithm in a Monte Carlo simulation.
Note that the proposed inverse formula does not require
iterative computations (except for matrix inversion),
while the gradient ascent algorithm requires thousands of iterations.

Although we have demonstrated that
the inverse formula we derived in tree-reweighted approximation is useful,
some open questions should be solved for future improvement and practical applications.
First, we have the free parameter $\rho_{ij}$
which was fixed to uniform values in our experiments.
There is no doubt that
optimizing $\rho_{ij}$ will improve the accuracy of
the inverse formula.
In fact, an optimal $\rho_{ij}$ value
that minimizes the upper bound
$\Phi^{\rm TRW} (\theta; \rho)$ has been discussed previously \cite{wainwright}.
However, it is difficult to obtain an optimal value
by solving equations analytically, and
we must perform iterative computations.
Even though it is difficult to obtain the optimal value analytically,
there may be an easy choice of $\rho_{ij}$ that is superior
to the uniform choice used in this study.

Another question is the extension of the inverse formula
to models with hidden variables, e.g. restricted Boltzmann machines.
The introduction of hidden variables makes
the model drastically simpler and recognizable to a human being.
We may extend the inverse formula directly to include hidden variables,
or we may use the inverse formula in a step of expectation-maximization-like algorithms
to reduce computational cost.

Finally, we are interested in applying
tree-reweighted approximation and the proposed inverse formula to physics.
Note that the partition function
dominates the physical properties of a system, such as phase transitions
and critical phenomena:
thus, the tree-reweighted approximation, which can give the rigorous
bound of the partition function, may play an important role
in mathematical analysis of physical models.
The proposed formulation, with which we analyzed the exact solution of
the tree-reweighted free energy,
may also be useful to give new insights into statistical and mathematical physics.


\section*{Acknowledgment}
The author is grateful to Yuuji Ichisugi and anonymous reviewers for comments.

This paper is based on results obtained from a project
commissioned by the New Energy and Industrial Technology
Development Organization (NEDO).



%

\bibliographystyle{IEEEtran}
\bstctlcite{IEEEexample:BSTcontrol}


\end{document}